\documentclass[runningheads]{llncs}
\usepackage{cite}
\usepackage[T1]{fontenc}
\usepackage{algorithm}
\usepackage{algpseudocode}
\usepackage{multirow}
\usepackage{mathrsfs}
\usepackage{amsmath}
\usepackage{graphicx}
\usepackage{color}
\usepackage{upgreek}
\usepackage{bbm}
\usepackage{bm}
\usepackage[utf8]{inputenc}
\usepackage{url}
\usepackage{microtype} 
\usepackage{amssymb} 
\definecolor{cb_orange}{rgb}{1.0,0.51,0.0}
\definecolor{cb_blue}{rgb}{0.22,0.49,0.72}
\definecolor{cb_green}{rgb}{0.3,0.67,0.29}
\definecolor{cb_red}{rgb}{0.89,0.1,0.11}
\definecolor{cb_pink}{rgb}{1, 0, 0.4}

\begin{document}
\title{Reference-Free Isotropic 3D EM Reconstruction using Diffusion Models}
%
%
%
\author{Kyungryun Lee \and Won-Ki Jeong}
%
%
\institute{Korea University, College of Informatics, \\
 Department of Computer Science and Engineering \\
\email{\{krlee0000, wkjeong\}@korea.ac.kr}}
\maketitle              
\begin{abstract}
\label{0_abstract}


Electron microscopy (EM) images exhibit anisotropic axial resolution due to the characteristics inherent to the imaging modality, presenting challenges in analysis and downstream tasks. 
%
%
Recently proposed deep-learning-based isotropic reconstruction methods have addressed this issue; however, training the deep neural networks require either isotropic ground truth volumes, prior knowledge of the degradation process, or point spread function (PSF).
Moreover, these methods struggle to generate realistic volumes when confronted with high scaling factors (e.g. $\times$8, $\times$10)
In this paper, we propose a diffusion-model-based framework that overcomes the limitations of requiring reference data or prior knowledge about the degradation process. 
Our approach utilizes 2D diffusion models to consistently reconstruct 3D volumes and is well-suited for highly downsampled data. 
Extensive experiments conducted on two public datasets demonstrate the robustness and superiority of leveraging the generative prior compared to supervised learning methods. 
Additionally, we demonstrate our method's feasibility for self-supervised reconstruction, which can restore a single anisotropic volume without any training data.
The source code is available on GitHub: \url{https://github.com/hvcl/diffusion-em-recon}.

\end{abstract}
\section{Introductions}
\label{1_introduction}
%
While 3D electron microscopy (EM) provide exceptional lateral resolution of 3 to 5 nanometers per pixel, the prevalent technique of 3D EM imaging involves physically sectioning tissue samples, resulting in a significantly lower axial resolution of approximately 30 to 50 nanometers per pixel (i.e., section thickness). 
This lower axial resolution poses challenges particularly for small structures such as synaptic clefts that can be smaller than the section thickness.~\cite{connectome}
Conventional approaches, such as interpolation and deconvolution, have been used to address this issue, offering fast solutions. 
However, these methods often produce unsatisfactory results, particularly when dealing with texture-rich EM images.

In recent years, deep learning-based methods have emerged as promising approaches for the 
isotropic reconstruction of EM images, outperforming the classical techniques.
Heinrich et al.~\cite{3d_em}
leveraged isotropic FIB-SEM images to generate training data for supervised training of a 3D UNet-based super-resolution model. 
%
%
However, acquiring isotropic data in real-world scenarios is not feasible and there are cases where the downsampling process is unknown. 
On the other hand, by leveraging the point spread function (PSF), several studies~\cite{fluroscence, content_aware} proposed a framework for fluorescence microscopy that does not require isotropic training data. 
By convolving the PSF with laterally viewed high-resolution image and subsampling, they simulated the anisotropic axial images.
%
Training a 2D-UNet-like architecture using the generated pairs, they achieved superior performance compared to conventional deconvolution algorithms\cite{RL}. 
An interesting aspect of such approaches is the potential for self-supervised learning, as the target data itself can be used for training.
%
Building upon this work, Deng et al.~\cite{degradation_learning} conducted further experiments under more realistic settings. 
They demonstrated that self-supervised training using an inaccurate PSF could yield poor results. 
To address the limitation, they adopted a cycle-GAN\cite{cycle_gan} framework to implicitly learn the degradation process and generate proper low-resolution versions for training.
%
%
Nevertheless, these works rely on deterministic reconstruction models that aim to minimize the pixel-wise error; hence, when performed on high-scaling factors ($\times$8, $\times$10), the results are blurry and fail to preserve fine structures.

In this study, to tackle a challenging scenario where no training data is available, we propose a novel approach that leverages the denoising diffusion probabilistic model (DDPM)~\cite{ddpm} for realistic 3D EM reconstruction. %
The diffusion model is recently gaining attention due to its high fidelity and diverse generation compared to other generative models~\cite{ddim, ddpm_vs_gan}. 
The diffusion model is adopted in various tasks for not only natural image domains~\cite{sr3, ilvr, palette} 
but also for medical image modalities~\cite{score_mri,med_diff,anomaly,3d2d_diff}. 
They are also well known for handling inverse problems\cite{ddrm, ddnm, posterior_sampling}.
%
In particular, these methods are capable of restoring 2D images without requiring task-specific training or datasets.
%
Consequently, considering that our 3D reconstruction problem can be regarded as a super-resolution (SR) task, which inherently is an inverse problem, we leverage diffusion models to address it.
%
%
However, in the context of 3D generation, it becomes necessary to employ a generative model that can capture the underlying 3D data.
%
%
Training such a model is challenging not only due to the significant memory resources required but also acquiring isotropic 3D training data is not feasible. 
Hence, we adopt a slice-by-slice approach by utilizing 2D diffusion models for the reconstruction of the 3D volume.
We train a 2D diffusion model to learn the data distribution from high-resolution lateral images. 
Subsequently, we leverage the diffusion prior of the laterally trained model in the sequential reconstruction of the low-resolution axial images. 
%
%
In order to maintain coherence across the 3D volume during the independent 2D reconstructions, we propose a sampling strategy where the previously reconstructed slice is encoded and used as a reference for the reconstruction of the next slice. 
We also introduce a heuristic method that improves the interpolated approximation to handle cases where the PSF is unknown, providing robustness in real-world scenarios.
%
%
We validate the effectiveness and stability of the two proposed strategies via experiments and ablation studies. 
%
%
%
Our main contributions can be summarized as follows:
\begin{itemize}        
    %
%
    \item We propose a sampling scheme that allows coherent 3D reconstruction using only 2D diffusion models. This allows smooth and continuous transitions between slices, therefore, eliminating artifacts when viewed in a perpendicular direction.
    \item We offer a simple but effective heuristic that can be applied without knowing the exact PSF which is often in practice. Moreover, the proposed approach is interpretable, allowing the reconstruction process to be more reliable.
    \item We demonstrate the superior reconstruction of DDPMs compared to previous auto-encoder-based methods by conducting simulation studies on a public dataset~\cite{fib25} for scenarios with/without prior information of the PSF. We also assess the performance on a real serial-section transmission electron microscopy (ssTEM)~\cite{cremi} volume without reference data or PSF information. 
    %
 %
     
\end{itemize}

\section{Method}
\label{2_methods}





\begin{figure}[t] 
    \centering
    \includegraphics[width=0.99\textwidth]{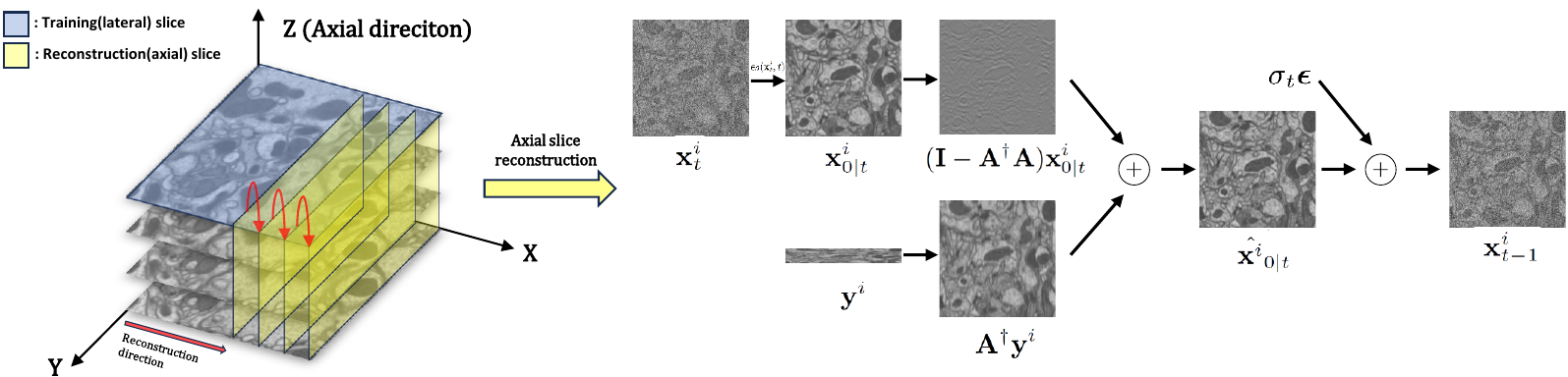}
    \caption{
    (left): Reconstruction strategy for 3D EM via 2D diffusion models.
    The lateral images are used for training a diffusion model. 
    Once trained, consistent sampling can be apllied for any kind of 2D degradation($\textbf{A}$).
    (right): Intuitive illustration of the refinement process. The low-frequency components are replaced with $\textbf{A}^\dagger\textbf{y}^i$ to fit the degradation process.
    }
    \label{fig:framework}
\end{figure}

As shown in Fig.~\ref{fig:framework}(left), the whole process can be divided into two steps. 
We initially train a 2D DDPM on the lateral images of our target volume. 
This allows our generative model to learn "How high resolution images look like".
Later, the laterally trained diffusion model is applied to reconstruct anisotropic axial planes slice-by-slice. 
%
Especially, we follow the restoring process of diffusion null-space model(DDNM)\cite{ddnm}.


\subsubsection{Preliminaries}
Diffusion models first define a $T$-step forward process that progressively perturbates an image to pure noise $\textbf{x}_T\sim{\mathcal{N}(0,\textbf{I})}$\cite{ddpm}. 
By defining the noise schedule parameters $\beta_t$, $\alpha_t:=1-\beta_t$ and $\bar{\alpha_t}:=\prod_{s=1}^{t}\alpha_s$, the forward process can be marginalized to a simple closed form of
\begin{equation}
    q({\textbf{x}_t}|\textbf{x}_0)=\mathcal{N}(\textbf{x}_t;\sqrt{\bar\alpha_t}\textbf{x}_0,(1-\bar\alpha_t)\textbf{I}),
    \label{eq:forward}
\end{equation}
The reverse process can be thought of as sampling from the posterior distribution,
$q(\textbf{x}_{t-1}|\textbf{x}_t, \textbf{x}_0).$
Therefore, to estimate the true posterior with $p_\theta(\textbf{x}_{t-1}|\textbf{x}_t)$, a noise predicting model is trained by minimizing the loss:
\begin{equation}
    \mathcal{L}_{t-1}=\mathbb{E}_{\textbf{x}_0,\bm{\epsilon},t}
    [||\bm{\epsilon}-\bm{\epsilon}_\theta(\sqrt{\alpha_t}\textbf{x}_0+\sqrt{1-\alpha_t}\bm{\epsilon},t)||_2]
    \label{eq:loss}
\end{equation}
As described in DDIM\cite{ddim}, reparameterization allows inference in two steps by estimating the noise with $\bm{\epsilon}_\theta$ as follows:
\begin{equation}
    \textbf{x}_{0|t} = \frac{1}{\sqrt{\bar\alpha_t}}(\textbf{x}_t - \bm{\epsilon}_\theta(\textbf{x}_t,t)\sqrt{1-\bar\alpha_t})
    \label{eq:x0t}
\end{equation}
\begin{equation}
    \textbf{x}_{t-1} = \sqrt{\bar\alpha_{t-1}}\textbf{x}_{0|t} + 
    \sqrt{1-\bar\alpha_t-\sigma_t^2} \cdot \bm{\epsilon}_\theta(\textbf{x}_t,t)
     + \sigma_t\bm{\epsilon},\quad \bm{\epsilon}\sim{\mathcal{N}(0,\textbf{I})},
     \label{eq:reverse}
\end{equation}
with $\sigma_t = \sqrt{(1-\bar\alpha_{t-1})/(1-\bar\alpha_t)}\sqrt{1-\bar\alpha_t/\bar\alpha_{t-1}}$.
Roughly speaking, at every iteration the reverse process is first estimating the clean image $\textbf{x}_{0|t}$ at time $t$ and again perturbing it with noise level $t-1$, gradually decreasing the noise until $t=0$. 

DDNM\cite{ddnm} builds upon this inference process to solve linear inverse problems that are generally defined as 
\textbf{y} = \textbf{Ax}, 
where we aim to restore the data $\textbf{x} \in \mathbb{R}^{n \times 1}$, given the degradation matrix $\textbf{A} \in \mathbb{R}^{m \times n}$ and its observation $\textbf{y} \in \mathbb{R}^{m \times 1}$.
To sample an image that fits the constrain given by $\textbf{A}$, 
range-space replacement is added 
after Eq.~\ref{eq:x0t} 
as follows:
\begin{equation}
    \hat{\textbf{x}}_{0|t} = \textbf{A}^\dagger\textbf{y} + (\textbf{I}-\textbf{A}^\dagger\textbf{A})\textbf{x}_{0|t}
    \label{eq:ddnm_refinement}
\end{equation}
where $\textbf{A}^\dagger \in \mathbb{R}^{m \times n}$ is the pseudo-inverse of $\textbf{A}$ which can be calculated by the singular value decomposition (SVD) method.
This refinement process ensures $\hat{\textbf{x}}_{0|t}$ to satisfy the linear condition, hence shifting the direction of the reverse process to be consistent with the degradation.
%
By substituting $\textbf{x}_{0|t}$ with $\hat{\textbf{x}}_{0|t}$ in Eq.~\ref{eq:reverse}, the noise mitigates the discrepancies between the replaced and original components of $\hat{\textbf{x}}_{0|t}$. Fig.~\ref{fig:framework}(right) illustrates the refinement process for isotropic reconstruction.
%
%
%
\subsubsection{Diffusion Models for 3D EM Reconstruction}
\label{sec:diffmodelfor3dem}
\begin{algorithm}[t]
\label{algorithm}
\caption{Reconstruction of the i\textsuperscript{th} slice}
\label{alg:alg1}
\begin{algorithmic}
\Require $\textbf{y}^i$, $\textbf{x}_0^{i-1}$, ${\textbf{A}}$, ${\textbf{A}^\dagger}$, $\bm{\epsilon}_\theta$
\For{$t = 0, ..., {R-1}$} \Comment{Encode $\textbf{x}_0^{i-1}$ deterministically}
    \State $\textbf{x}_{0|t}^{i-1} = \frac{1}{\sqrt{\bar\alpha_t}}(\textbf{x}_t^{i-1} -         
            \bm{\epsilon}_\theta(\textbf{x}_t^{i-1},t)\sqrt{1-\bar\alpha_t})$    
    \State $\textbf{x}_{t+1}^{i-1} = \sqrt{\bar\alpha_{t+1}}\textbf{x}_{0|t}^{i-1} + 
            \sqrt{1-\bar\alpha_{t+1}} \cdot \bm{\epsilon}_\theta(\textbf{x}_t^{i-1},t)$
\EndFor 
\State $\textbf{x}_R^{i}$ = $\textbf{x}_R^{i-1}$
\For{$t = R, ..., 1$} \Comment{DDNM reconstruction starting from $\textbf{x}_R^{i-1}$}
    \State $\bm{\epsilon}\sim{\mathcal{N}(0,\textbf{I})}$
    \State $\textbf{x}_{0|t}^i = \frac{1}{\sqrt{\bar\alpha_t}}(\textbf{x}_t^i -
            \bm{\epsilon}_\theta(\textbf{x}_t^i,t)\sqrt{1-\bar\alpha_t})$    
    \State $\hat{\textbf{x}^i}_{0|t} = \textbf{A}^\dagger\textbf{y}^i   
            + (\textbf{I}-\textbf{A}^\dagger\textbf{A})\textbf{x}_{0|t}^i$
    \State $\textbf{x}_{t-1}^i = \sqrt{\bar\alpha_{t-1}}\hat{\textbf{x}}_{0|t}^i + 
            \sqrt{1-\bar\alpha_t-\sigma_t^2} \cdot \bm{\epsilon}_\theta(\textbf{x}_t^i,t)
            + \sigma_t\bm{\epsilon} $
\EndFor
\end{algorithmic}
\end{algorithm}

%
As the degradation process occurs along the Z-axis of the 3D volume, it can be simplified as 2D degradations of contiguous ZY(or ZX) images. 
The 2D degradation can be represented with a matrix
$\textbf{A} = \textbf{S}_f\textbf{P}$, 
where  $\textbf{S}_f \in \mathbb{R}^{m \times n}$ is the sub-sampling operator choosing every $f$ rows and 
$\textbf{P} \in \mathbb{R}^{n \times n}$ is the PSF convolution operator.
Assuming that we know the PSF and the downsampling factor, we can construct $\textbf{A}$ and its pseudo-inverse $\textbf{A}^\dagger$, therefore we directly apply the DDNM sampling procedure to the i\textsuperscript{th} low-resolution ZY slice $\textbf{y}^i$ to reconstruct $\textbf{x}^i$.
%
However, it is important to regard that the diffusion model used in the 2D reconstruction does not take into account the continuity 
between neighboring slices.
%

%
%
Therefore, we propose a consistent sampling strategy where the previous slice is encoded by DDIM and used as a starting point for the subsequent slice generation.
This approach brings continuity 
between neighboring slices by leveraging the information encoded in the preceding slice. 
%
Moreover, most of the information overlaps between neighboring slices, therefore referencing the previous slice eases the reconstruction process for the diffusion model.
Rather than beginning with pure Gaussian noise, we start the generation process of image $\textbf{x}^i$ by encoding the previously reconstructed images $\textbf{x}^{i-1}_0$ in a sequence of $[1, ..., R]$ and use $\textbf{x}^{i-1}_R$ as a starting point.
Specifically, by setting $\sigma_t=0$ in Eq.~\ref{eq:reverse} 
the DDIM iteration loses its stochasticity and it is possible to encode an image through a deterministic forward process.
Given $\textbf{x}^{i-1}_R$ and $\textbf{y}^i$, the i\textsuperscript{th} slice is reconstructed by the reverse process with $[{R-1}, ...,0]$, but this time with the introduction of random noise.
As there is no previous reference for the first slice, the reverse diffusion process starts from Gaussian noise($R=1000$).
The overall process is described in Algorithm~\ref{alg:alg1}.
We also observed that, although our method allows smooth transition along the sampling axis, the perpendicular planes do not directly leverage the diffusion prior, thus showing unrealistic visual results.
Therefore, we ensemble the two reconstructions processed along the x-axis and y-axis. 
Additionally, due to the generative model's stochastic nature, averaging the two results show a more steady and reliable generation.
%
%

In certain scenarios, the exact PSF is unknown.
%
%
Therefore, we propose a simple approximation where we set $\textbf{A}$ as linear down-sampling and $\textbf{A}^\dagger$ as a linear interpolation operator. 
Despite the fact that linear interpolation is not the exact pseudo-inverse of linear down-sampling, 
\cite{ilvr} adopts it as a low-frequency guidance to generate an image in a desired direction.
This approach enables the diffusion model to fill in the missing high-frequency details on top of the blurry interpolated observation $\textbf{A}^\dagger\textbf{y}$. 
As a result, the reconstructed data preserves the low-frequency structural information of the interpolated observation and remains interpretable without introducing abrupt changes.
Although it has limitations that the reconstruction is a heuristic that relies on interpolation, we demonstrate through ablation that it gives better results compared to other assumptions of $\textbf{A}$. 
We note that other kinds of interpolation methods can be used for $\textbf{A}$ and $\textbf{A}^\dagger$ instead of linear, e.g. cubic or lanczos.

\section{Experiments}
\label{3_experiments}
We assess the performance of our framework using two widely-used EM datasets: FIB-25~\cite{fib25} and CREMI~\cite{cremi}. 
%
FIB-25 dataset is an isotropic FIB-SEM data commonly used for simulation studies, allowing quantitative evaluation of performance.
%
CREMI is a ssTEM dataset with an anisotropic axial resolution. It serves as a real-world dataset for evaluating the performance of algorithms in handling anisotropic data.
We trained the diffusion model with a U-Net backbone following\cite{ddpm} and adapted cosine scheduling\cite{iddpm} where $T=1000$. 
The lateral training image size is $512\times512$ and the batch size is 4 with a learning rate of 0.00002.
For sequential sampling, we reconstruct ZY images slice-by-slice along the x-axis, where the encoding/decoding level is $R=200$.
Except for the first axial slice, all slices were encoded by 4 steps and reconstructed(decoding) by 50 steps, where strided steps allow faster sampling.
We compare our method with three auto-encoder-based approaches including 3D-SR-UNet\cite{3d_em}, IsoNet\cite{fluroscence, content_aware} and the framework proposed by Deng et al.\cite{degradation_learning}.
All the methods were implemented in Pytorch\cite{pytorch} and tested on a single NVIDIA RTX A6000 GPU.

%

%
\subsubsection{Evaluation on simulated data}
A randomly chosen subvolume of size $512\times512\times512$ from the isotropic FIB-25 data is convolved with a Gaussian filter and downsampled by choosing every $f \in \{4, 8\}$ lateral slices throughout the Z direction to generate synthetic anisotropic data.
All deep-learning approaches except 3D-SR-UNet use only the target volume itself for training.
%
3D-SR-UNet is trained in a fully supervised manner with 26 additional isotropic subvolumes.
%
%
%
Experiments are conducted for both cases where we know the PSF or not.
For 3D-SR-UNet and IsoNet, we use average downsampling to generate low-resolution pair for blind-PSF. 
%
%
All methods require separate training for different PSF and downsampling factors, whereas our method does not require additional training.
Performance is quantitatively evaluated by not only PSNR but also multi-scale structural similarity (MS-SSIM)~\cite{ms-ssim} and LPIPS~\cite{lpips}, which are more sensitive to fine patterns and structural details to demonstrate the realistic reconstruction achieved by our approach.

\begin{table}[tb]
\centering
\caption{Quantitative Comparison with other methods. With $f=8$ and Gaussian filter of $\sigma=4$, the isotropic FIB25 volume is simulated to an anisotropic resolution of $64\times512\times512$ The baseline is linear interpolation. $\perp$ and $+$ indicate that the reconstruction was done on ZX slices or is ensembled, respectively.The highest scores are highlighted in \textbf{bold}.}
\label{table:comp}
\renewcommand{\arraystretch}{1.0}
\setlength{\tabcolsep}{0.3pt}
\scriptsize{
\begin{tabular}{|c|c|ccc|ccc|ccc|}
\hline
\multirow{3}{*}{\textbf{PSF}} & \multirow{2}{*}{\textbf{Method}} & \multicolumn{3}{|c|}{\textbf{ZY}}  &\multicolumn{3}{|c|}{\textbf{ZX}} & \multicolumn{3}{|c|}{\textbf{XY}} \\
& & PSNR$\uparrow$ & MS-SSIM$\uparrow$ & LPIPS$\downarrow$ & PSNR$\uparrow$ & MS-SSIM$\uparrow$ & LPIPS$\downarrow$ & PSNR$\uparrow$ & MS-SSIM$\uparrow$ & LPIPS$\downarrow$ \\
\cline{2-11}
& Baseline & 26.12 & 0.842 & 0.567 & 26.11 & 0.840 & 0.555 & 26.18 & 0.848 & 0.379 \\
\cline{1-11}

\multirow{4}{*}{Exact} & 3D-SR-UNet\cite{3d_em} & \textbf{28.96} & \textbf{0.934} & 0.486 & \textbf{28.97} & \textbf{0.931} & 0.479 & \textbf{29.04} & \textbf{0.931} & 0.412\\
\cline{2-11}
& IsoNet\cite{fluroscence} & 28.62 & 0.928 & 0.490 & 28.56 & 0.924 & 0.495 & 28.67 & 0.922 & 0.328 \\
\cline{2-11}
& Ours & 27.92 & 0.914 & \textbf{0.375} & 27.93 & 0.916 & 0.434 & 28.03 & 0.913 & 0.296 \\
\cline{2-11}
& Ours\textsuperscript{$+$} & 28.39 & 0.924 & 0.426 & 28.39 & 0.922 & \textbf{0.425} & 28.49 & 0.920 & \textbf{0.264} \\
\hline

\multirow{6}{*}{Blind} & 3D-SR-UNet & 27.52 & 0.894 & 0.512 & 27.52 & 0.891 & 0.503 & 27.57 & 0.895 & 0.426 \\
\cline{2-11}
& IsoNet & 27.60 & 0.897 & 0.503 & 27.29 & 0.888 & 0.515 & 27.35 & 0.889 & 0.363 \\
\cline{2-11}
& Deng \textit{et al.}\cite{degradation_learning} & 27.65 & 0.901 & 0.496 & 27.65 & 0.901 & 0.504 & 27.75 & 0.900 & 0.408 \\
\cline{2-11}
& Ours & 27.55 & 0.901 & \textbf{0.391} & 27.55 & 0.903 & 0.448 & 27.64 & 0.901 & 0.325 \\
\cline{2-11}
& Ours\textsuperscript{$\perp$} & 27.57 & 0.905 & 0.453 & 27.57 & 0.900 & \textbf{0.393} & 27.66 & 0.901 & \textbf{0.280} \\
\cline{2-11}
& Ours\textsuperscript{$+$} & \textbf{27.95} & \textbf{0.911} & 0.431 & \textbf{27.95} & \textbf{0.909} & 0.433 & \textbf{28.04} & \textbf{0.908} & 0.284 \\
\hline
\end{tabular}
}
\end{table}

\begin{figure}[t]
    \centering
    \includegraphics[width=0.99\textwidth]{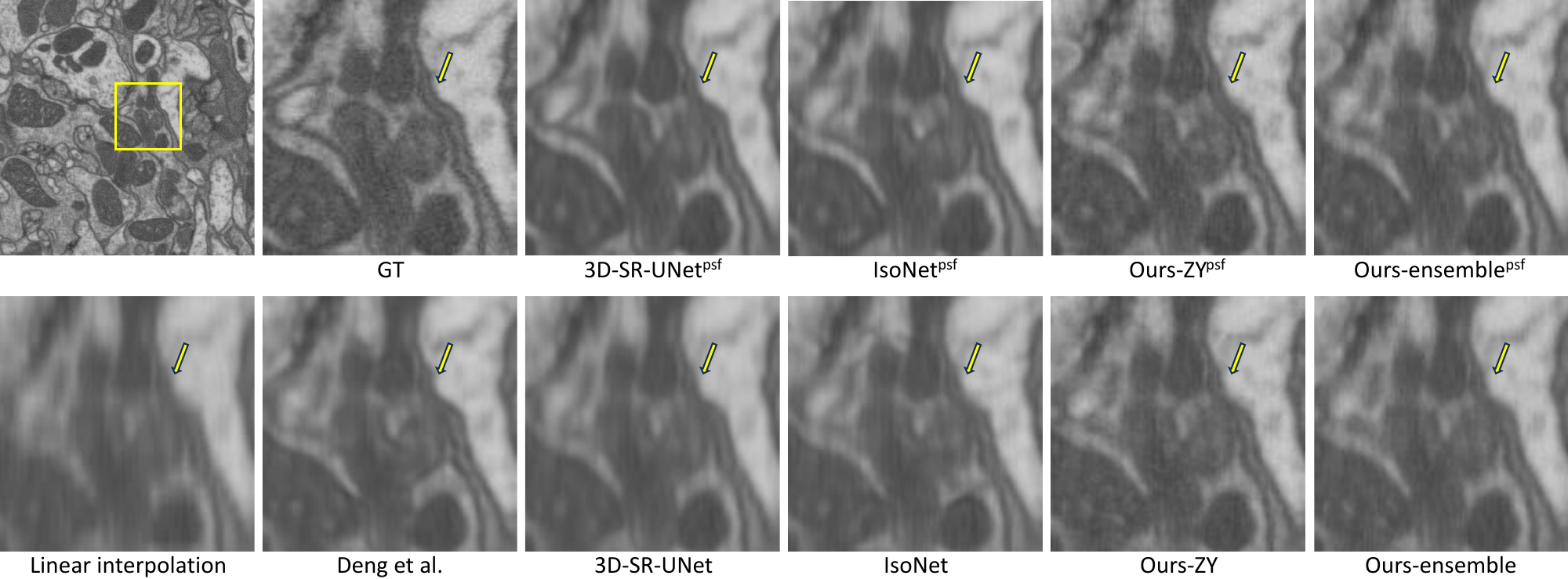}
    \caption{
    Qualitative comparison of FIB25 reconstruction viewed in ZY. $f=8$ and a Gaussian filter of $\sigma=4$ was used. 
    The superscript "psf" indicates that the exact point spread function was used for reconstruction.
    }
    \label{figure:fib25}
\end{figure}

For $\sigma=4$ and $f=8$,
Table~\ref{table:comp} shows that 
our method outperforms all approaches for blind-PSF scenarios, especially in terms of LPIPS which measures the perceptual similarities between the reference and reconstructed images. 
%
%
In the case where we know the exact PSF, 3D-SR-UNet and IsoNet show better performance in PSRN and MS-SSIM.
Nevertheless, the LPIPS score and visual results confirm that they cannot generate high-quality results.
Moreover, 3D-SR-UNet is trained with isotropic volumes, thus incomparable.
As discussed in Section~\ref{sec:diffmodelfor3dem}, `Ours\textsuperscript{$\perp$}' shows that reconstruction along the ZX planes fails to generate realistic images viewed in ZY. Although the pixel-wise metrics preserve, the LPIPS score drops dramatically.
`Ours\textsuperscript{$+$}' averages the two reconstructions 
along ZX and ZY, yielding a compromised result viewed in all directions and resulting in higher PSNR and MS-SSIM scores. 
%
%
Fig.~\ref{figure:fib25} shows that the auto-encoder-based approaches tend to produce blurry results. 
This may be due to the limitation of the deterministic models based on pixel-wise loss functions, which may not fully capture the complexity and intricate details of the data. 
Furthermore, a noticeable quality gap exists between blind-PSF and exact-PSF cases for all methods except ours, which supports the robustness of our proposed heuristic.
Experimental results for $\sigma=2$ and $f=4$ are described in Table S1 and Fig S1.
%
%
%
%


\subsubsection{Visual comparison with real ssTEM data}
\begin{figure}[h]
    \centering
    \includegraphics[width=0.90\textwidth]{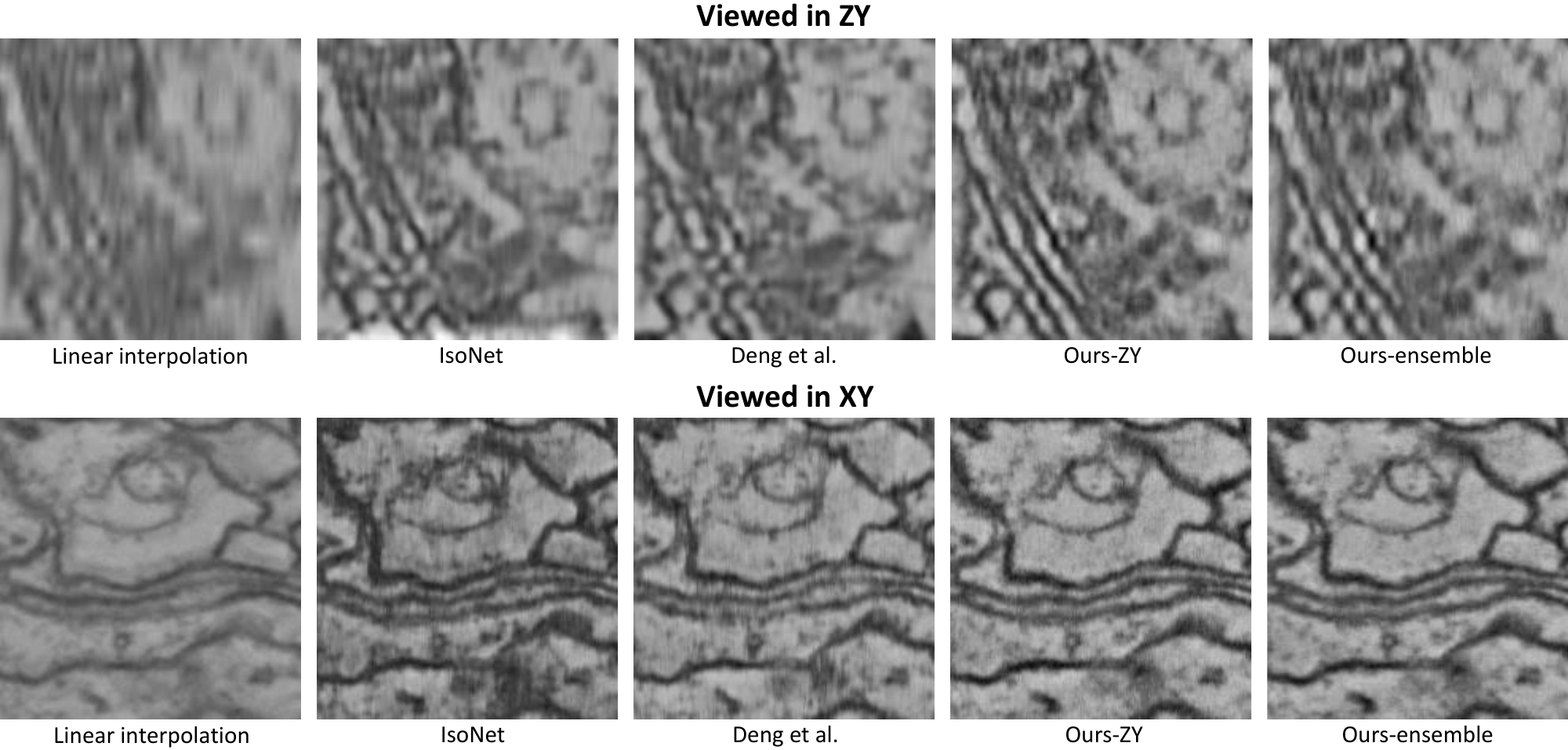}
    \caption{
    Visual results of the reconstruction of a CREMI volume viewed in ZY and XY.
    }
    \label{figure:cremi}
\end{figure}

In this section, we present reconstruction results of real anisotropic ssTEM data (CREMI) from $52\times512\times512$ to $512\times512\times512$, which does not have PSF information nor isotropic reference data. 
Thus, 3D-SR-UNet cannot be trained.
For IsoNet, we use average downsampling to generate the low-resolution pairs for training.
Our method uses the interpolation guidance method introduced in Section~\ref{sec:diffmodelfor3dem} for reconstruction.
Fig.~\ref{figure:cremi} shows the reconstructed ssTEM volumes viewed in ZY and XY.
IsoNet and Deng et al.'s method 
show blurry results for ZY. Moreover, the XY images show severe artifacts indicating misalignment. 
%


\subsubsection{Ablation studies}
\begin{figure}[h]
    \centering
    \includegraphics[width=0.90\textwidth]{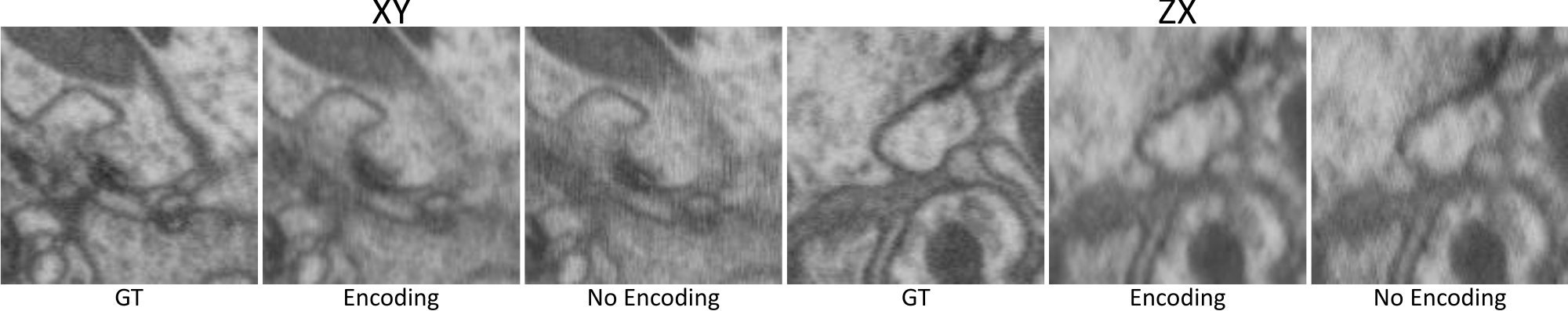}
    \caption{
    Blind-PSF reconstructed volume viewed in XY and ZX. Reconstruction without referencing the previous slice shows severe artifacts due to the misalignment of adjacent slices. 
    }
    \label{figure:ablation}
\end{figure}

We perform two ablation studies on the FIB-25 dataset anisotropically simulated with $\sigma=4$ and $f=8$, starting with a comparison of different assumptions for the degradation process in blind-PSF scenarios.
Imputation refers to the direct filling in of missing information, similar to the process of inpainting.
We also compare it with average downsampling and an incorrect Gaussian filter of $\sigma=2$ (note that $\sigma=4$ is used to generate the synthetic anisotropic data).
Table S2 and Fig S2 suggests that reconstructing on top of the interpolated approximation gives better results than imputing or estimating the degradation using an incorrect filter. 
%
%
Secondly, we investigate the importance of continuous sampling throughout the reconstruction process. 
As shown in Fig.~\ref{figure:ablation}, when the previous slice is not encoded as a reference, the reconstruction exhibits visible artifacts in XY and ZX views.
%
%

\section{Conclusion}
\label{4_conclusion}
We present a diffusion-model-based approach for reference-free isotropic reconstruction on highly anisotropic 3D EM volumes.
We introduced two additional strategies that allow 2D diffusion models for consistent 
3D reconstruction where the PSF is unknown.
Through quantitative and qualitative results, we demonstrated its superiority compared to SOTA reconstruction methods and showed the limitations of auto-encoder-based frameworks.
In addition to generating high-quality data, it exhibits efficacy in challenging conditions where training data is scarce and prior information is minimal, thereby demonstrating its potential in real-world applications.
In the future, we plan to investigate the impact of our methods on various downstream tasks in biomedical domains.

\subsection*{\textbf{Acknowledgements}}This work was partially supported by the National Research Foundation of Korea (NRF-2019M3E5D2A01063819, NRF-2021R1A6A1A
13044830), the Institute for Information \& Communications Technology Planning
\& Evaluation (IITP-2023-2020-0-01819), the Korea Health Industry Development
Institute (HI18C0316), the Korea Institute of Science and Technology (KIST)
Institutional Program (2E32210 and 2E32211), and a Korea University Grant.

\bibliographystyle{splncs04}
\bibliography{5_reference}

\clearpage


\end{document}